%% file: main.tex
\documentclass[10pt,twocolumn,letterpaper]{article}

\usepackage[pagenumbers]{wacv} % To force page numbers, e.g. for an arXiv version

\input{preamble}

\definecolor{wacvblue}{rgb}{0.21,0.49,0.74}
\usepackage[pagebackref,breaklinks,colorlinks,allcolors=wacvblue]{hyperref}

\def\wacvPaperID{***} % *** Enter the WACV Paper ID here
\def\confName{WACV}
\def\confYear{2027}

\title{Adapt, Label, Distill: Foundation-Model Supervision for Efficient Object Detection}
\author{
Zhengye Yang\thanks{Equal contribution.} \quad
Anthony Wang\footnotemark[1] \quad
Yuyang Li \\
Mehmet Kerem Turkcan \quad
Beizhan Yan \quad
Zoran Kostic \\
Columbia University\\
New York, NY, USA\\
{\tt\small zy2318@columbia.edu}
}

\begin{document}
\maketitle
\input{sec/0_abstract}    
\input{sec/1_introd}
\input{sec/2_related_work}
\input{sec/3_method}

\input{sec/4_experiments}

\input{sec/5_analysis}

\input{sec/6_conclusion}
{
    \small
    \bibliographystyle{ieeenat_fullname}
    \bibliography{main}
}
\input{sec/X_supp}

\end{document}

%% file: preamble.tex
\usepackage{wrapfig}
\usepackage{pifont}
\usepackage{multirow}
\usepackage{colortbl}
\usepackage{algorithm}
\usepackage{algpseudocode}
\usepackage{amsmath,amssymb}
\usepackage[accsupp]{axessibility}  % Improves PDF readability for those with disabilities.

\newcommand{\oracle}[1]{\textcolor{gray}{#1}}

\def\maketitlesupplementary{
    \clearpage
    \twocolumn[
        \centering
        {\Large\bfseries \thetitle\par}
        \vspace{0.5em}
        {\Large Supplementary Material\par}
        \vspace{1.0em}
    ]
}

\definecolor{checkmark}{HTML}{305AFF}
\definecolor{xmark}{HTML}{E62020}

\definecolor{ours}{rgb}{1.0, 0.93, 0.78}

%% file: sec/0_abstract.tex
\begin{abstract}
Object detectors often transfer poorly to specialized domains,
where box annotations are scarce and deployment requires
compact, real-time models. Large vision foundation models
can provide offline supervision, but their zero-shot predictions
may be poorly aligned with the target domain.
We present an adapt--label--distill pipeline that uses limited
target annotations to specialize a foundation teacher and
transfer its supervision to compact detectors.
We adapt SAM3 on the labeled subset, generate pseudo-labels
for the remaining unlabeled images, and train YOLOv11 students
with confidence-guided supervision.
Teacher confidence weights pseudo-label losses, while
zero-gating retains uncertain hypotheses during assignment
without applying their associated training losses.
Experiments across six target domains and three student
capacities show substantial gains in several domains, while
revealing limited or negative transfer in others.
Teacher comparisons under a common hard-label training recipe
support the choice of SAM3, and comparisons with an
EfficientTeacher adaptation provide additional evidence
against an online SSOD baseline.
With YOLOv11m, our pipeline improves over STAC-style
self-training by up to 15.3 AP and reaches 51.6 AP on Garbage,
compared with 54.1 under full supervision.
On NVIDIA Jetson AGX Orin, the deployed student achieves
188 FPS versus 2.9 FPS for SAM3, benefiting from foundation-model
supervision without incurring the teacher's deployment cost. Code is available \href{https://github.com/zk2172-columbia/ProjectAdaptLabelDistill_ObjectDetection}{here}.
\end{abstract}

%% file: sec/1_introd.tex
% Revised, pipeline-oriented Introduction.
% Citation keys and teaser figure environment are taken from the original
% Introduction supplied by the author. The pipeline-figure label was not
% present in that source excerpt, so no unverified cross-reference is used.

\section{Introduction}
\label{sec:introduction}

\begin{figure}[t]
    \centering
    \includegraphics[width=\linewidth]{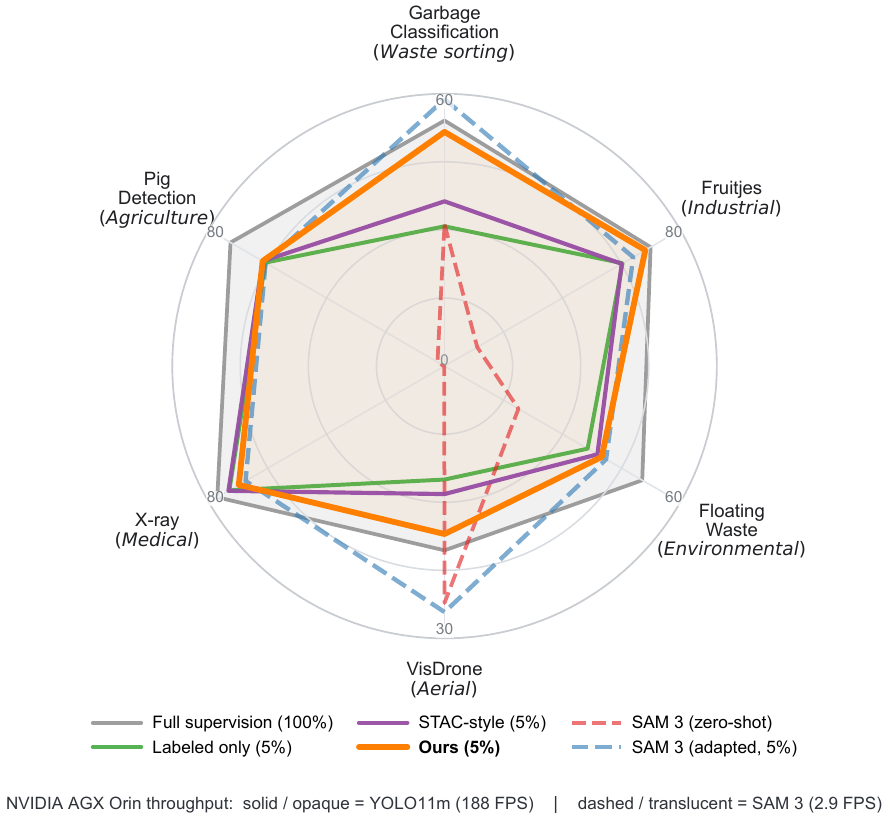}
    \caption{\textbf{Cross-domain accuracy and deployment efficiency.}
Each spoke reports mAP$_{50:95}$ using the dataset-specific scale shown at its
boundary. Our method and the STAC-style baseline use only $5\%$ labeled target
data, while the gray polygon denotes fully supervised YOLOv11m trained with
$100\%$ labels. The adapted SAM~3 teacher uses the same $5\%$ labeled subset.
Solid curves denote deployable YOLOv11m students (188 FPS), whereas dashed
curves denote the offline SAM~3 teacher (2.9 FPS), measured on NVIDIA AGX
Orin.}
    \label{fig:teaser}
    \vspace{-0.4cm}
\end{figure}

Object detectors are widely deployed in applications ranging from industrial inspection and agriculture to environmental monitoring and robotic systems. Yet, a detector that performs well on generic training data may struggle in a specialized environment, where appearance, viewpoints, and category definitions differ. Collecting bounding-box annotations for every new domain is costly, and practical deployments often require compact models that run in real time. Meanwhile, unlabeled target-domain images are comparatively easy to obtain. We study how to make effective use of these images and a limited annotation budget while keeping the deployed detector efficient.

Semi-supervised object detection (SSOD) uses pseudo-labels to learn from unlabeled images~\cite{sohn2020stac,liu2021unbiased,xu2021softteacher}. Existing methods have improved teacher--student consistency and pseudo-label assignment~\cite{chen2022dsl,wang2023consistentteacher,xu2023efficientteacher}, but typically derive their target-domain supervision from detectors initialized with the same limited annotations. Errors in those detectors, such as missed objects or incorrect boxes, can therefore propagate through self-training. This motivates exploring a complementary source of supervision.

Vision foundation models provide broad pretrained representations and can localize prompted concepts~\cite{li2022glip,liu2024groundingdino,carion2026sam3}. However, zero-shot predictions may not align with a specialized domain or its annotation policy. Rather than using scarce labels only to train the deployment detector, we explore using them first to adapt a foundation model into a domain-specific annotator. The adapted teacher can then generate supervision for a much larger unlabeled set, transferring its knowledge to a compact detector without transferring its inference cost.

We implement this strategy as an \emph{adapt--label--distill} pipeline. SAM~3~\cite{carion2026sam3} is adapted using the labeled target subset and then applied offline to produce pseudo-labels for the remaining images. We train YOLOv11 students with these annotations, using confidence-weighted losses and zero-gating to reduce the influence of uncertain predictions and potentially avoid treating uncertain object hypotheses as background~\cite{yang2020pu}. These mechanisms modify training only; the foundation teacher is not used at deployment.

We evaluate the pipeline across six target domains and three student capacities under a nominal $5\%$ annotation budget. As summarized in \Cref{fig:teaser}, foundation-model supervision improves student accuracy in several domains, although transfer is limited or negative in others. On Garbage Classification, YOLOv11m improves from 36.3 AP with STAC-style self-training to 51.6 AP, compared with 54.1 AP under full supervision. A stepwise ablation indicates that teacher adaptation accounts for most of the gain, with smaller contributions from confidence-aware training. On NVIDIA Jetson AGX Orin, the deployed YOLOv11m runs at 188 FPS at $640\times640$ resolution, versus 2.9 FPS for the offline SAM~3 teacher at its evaluated resolution.

Our contributions are threefold:
\begin{itemize}
    \item We study an \emph{adapt--label--distill} pipeline that uses limited target annotations to adapt a foundation model into an offline annotation teacher for efficient detectors.
    \item We evaluate external foundation-model supervision across six specialized domains, comparing pseudo-label teachers and SSOD baselines and documenting both positive and negative transfer.
    \item We validate deployment efficiency on NVIDIA Jetson AGX Orin, showing that the resulting compact students incur no additional inference-time teacher cost.
\end{itemize}

%% file: sec/2_related_work.tex
\section{Related Work}
\label{sec:related_work}

\noindent\textbf{Semi-Supervised Object Detection.}
Semi-supervised object detection (SSOD) exploits unlabeled images through
pseudo-labeling and teacher--student training. Early approaches such as
STAC~\cite{sohn2020stac} generate pseudo-boxes using a detector trained on the
labeled subset, while later methods maintain an online teacher and improve
pseudo-label quality through bias correction, confidence weighting, and
localization uncertainty~\cite{liu2021unbiased,xu2021softteacher,liu2022unbiasedv2}.
Other works improve dense pseudo-targets, foreground--background assignment, or
localization-aware selection~\cite{chen2022dsl,zhou2022denseteacher,li2022pseco,
wang2023consistentteacher}, including methods tailored to one-stage
detectors~\cite{xu2023efficientteacher}. Despite these differences, the
pseudo-label generator is itself learned primarily from the same limited
target annotations as the student. In contrast, we use a separately adapted
vision foundation model as an external annotation teacher. The teacher is used
only offline to annotate the unlabeled target set and is absent from the
deployed detector.

\noindent\textbf{Foundation Models as Annotation Teachers.}
Large-scale vision--language pre-training has enabled open-vocabulary detection
and grounding beyond a fixed training label space
~\cite{li2022glip,minderer2022owlvit,zhou2022detic,liu2024groundingdino}.
Promptable segmentation models provide complementary instance-level spatial
information: SAM~\cite{kirillov2023sam} generates masks from spatial prompts,
while SAM~3~\cite{carion2026sam3} additionally supports textual prompts and
returns matching object instances with masks and confidence scores. Such models
can therefore serve as automatic annotators, but their zero-shot predictions
need not match the appearance distribution, class definitions, or annotation
policy of a target dataset. We study both zero-shot and target-adapted SAM~3,
using the same limited target annotations available to the compact detector.

Our approach is also related to data and knowledge distillation. Data
Distillation~\cite{radosavovic2018datadistillation} converts model predictions
on unlabeled images into additional training annotations, whereas detection
distillation commonly transfers information through selected instances,
intermediate features, or localization distributions
~\cite{dai2021gid,yang2022fgd,zheng2022localizationdistillation}. We adopt an
annotation-based form of distillation: the foundation teacher and compact
student share neither parameters nor features, allowing the teacher's knowledge
to be transferred without inheriting its inference cost.

\noindent\textbf{Learning from Imperfect Pseudo-Labels.}
Pseudo-label quality varies substantially across detections, and confidence is
therefore commonly used to decide how strongly a prediction should supervise
the student. Standard SSOD typically retains predictions above a hard
confidence threshold~\cite{sohn2020stac,liu2021unbiased}, while later methods
use confidence or uncertainty more gradually through loss weighting, region
weighting, soft targets, or ignore assignments
~\cite{xu2021softteacher,wang2021datauncertainty,xu2023efficientteacher,
chen2022dsl}. Our confidence weighting follows this general principle at the
individual pseudo-box level: each matched pseudo-label contributes to the
student loss in proportion to the teacher's confidence.

Hard thresholding, however, creates a distinct problem when an uncertain
prediction corresponds to a real object. Discarding the pseudo-label removes
its positive supervision but leaves the object in the image, allowing its
region to be subsequently optimized as background. Positive-unlabeled and
sparse-label detection methods address related missing-annotation errors through
modified losses or proposal assignment
~\cite{yang2020pu,kaul2022labelverifycorrect,suri2023sparsedet}, while other
methods explicitly exploit low-confidence detections rather than discarding
them~\cite{liu2023lsm,yoon2024lpld}. Our \emph{zero-gating} mechanism follows
the same motivation but operates directly within pseudo-label assignment:
uncertain object hypotheses are retained so that they are not treated as
background, while their loss weight is set to zero. Thus, they provide neither
positive nor negative supervision.

Masking and erasing provide an alternative way to suppress uncertain visual
evidence~\cite{kim2022mum,li2022diverselearner,hoyer2023mic,fang2023upc}.
Unlike these image-space treatments, zero-gating leaves the training image
unchanged and requires only a change to the pseudo-label weight, introducing no
additional model components or deployment-time cost.

%% file: sec/3_method.tex
\section{Methodology}
\label{sec:method}

\begin{figure*}[t]
  \centering
  % \fbox{\rule{0pt}{2.2in} \rule{0.95\linewidth}{0pt}}
  \vspace{-0.6cm}
  \includegraphics[width=\linewidth]{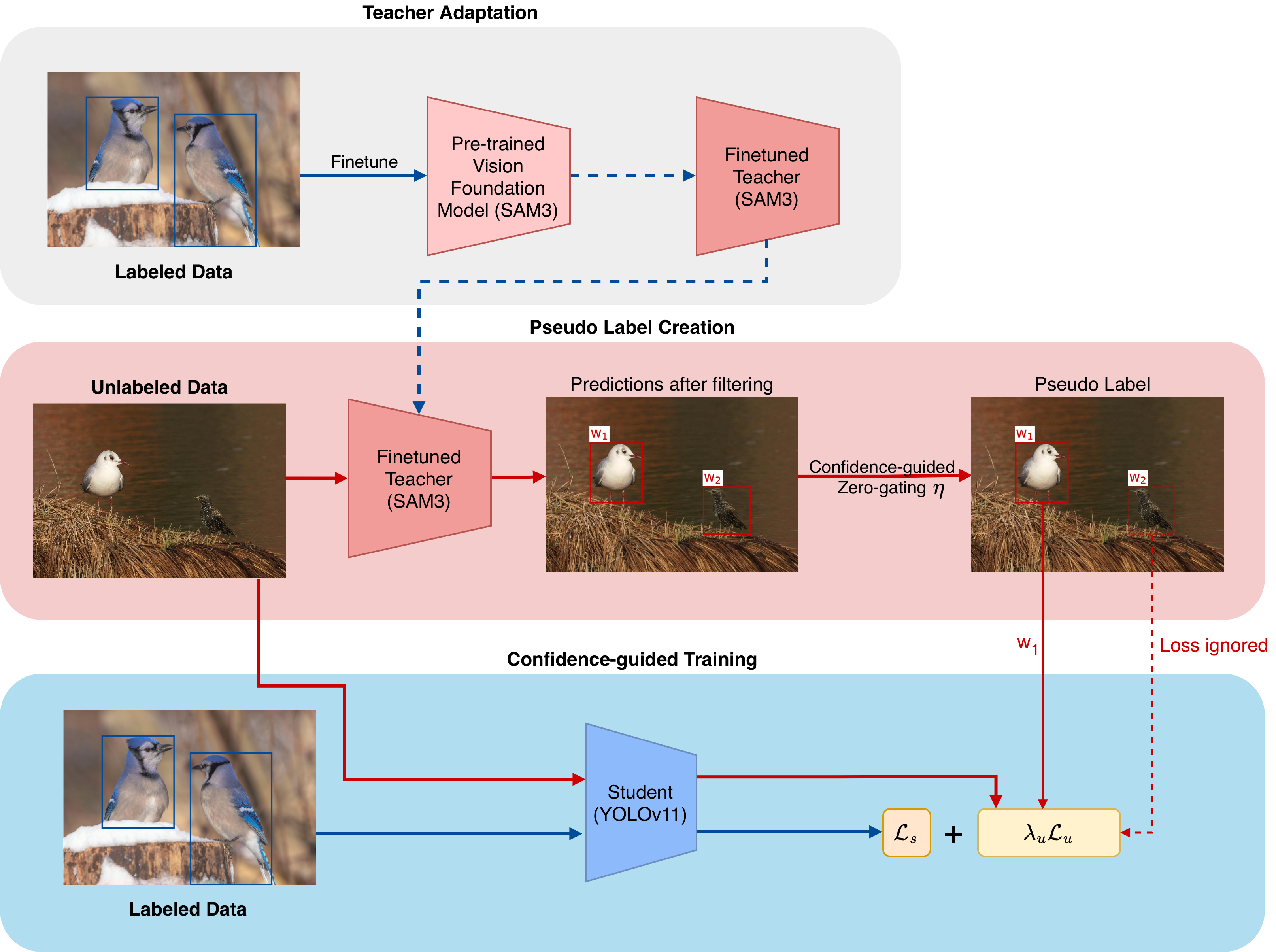}
  \caption{\textbf{Overview of the proposed pipeline.}
  A vision foundation model is fine-tuned on the small labeled set
  $\mathcal{D}_l$ and then applied once, offline, to the unlabeled set
  $\mathcal{D}_u$. Its raw predictions are filtered by confidence thresholding
  and non-maximum suppression, and the confidence score of every surviving
  detection is retained alongside its box and class. The compact student is
  trained on $\mathcal{D}_l$ with ground-truth annotations and on
  $\mathcal{D}_u$ with pseudo-labels, where each pseudo-label contributes to the
  loss in proportion to the teacher's confidence in it and uncertain
  pseudo-labels contribute nothing. The teacher is used only to produce the
  annotation set and is absent at deployment.}
  \label{fig:pipeline}
  \vspace{-0.4cm}
\end{figure*}

\begin{figure}[ht]
    \centering
    \includegraphics[width=\linewidth]{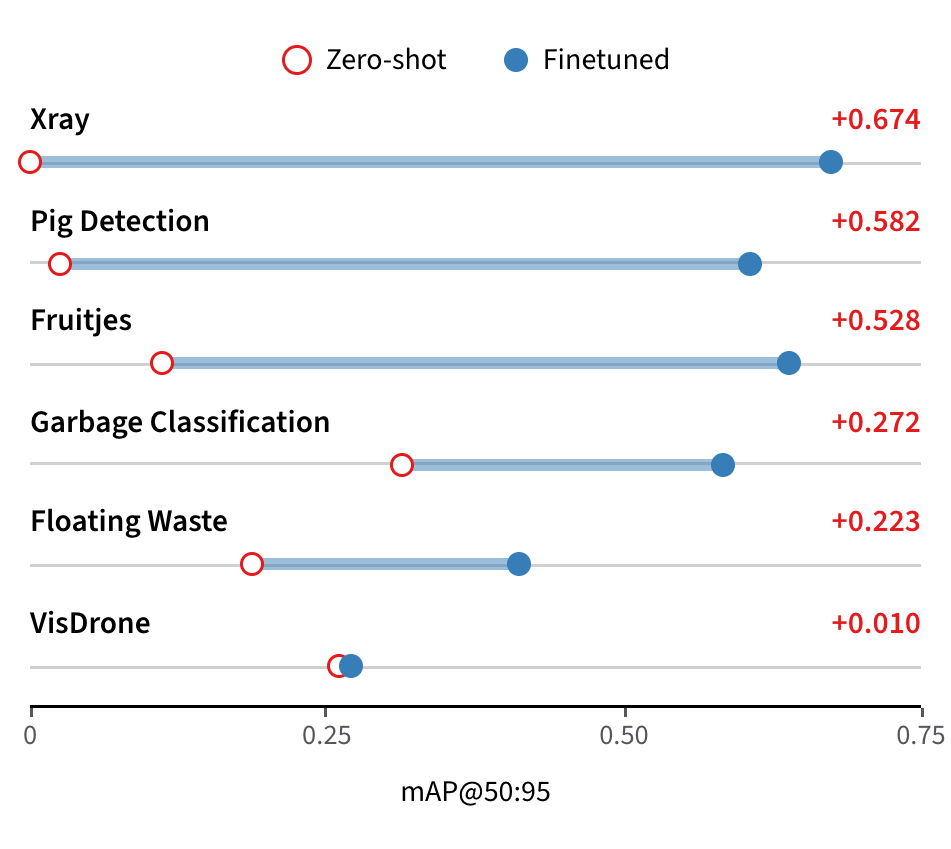}
    \vspace{-1.cm}
    \caption{\textbf{Effect of target-domain teacher adaptation.}
Performance of zero-shot and adapted SAM~3 on the test split of each target dataset. The teacher is adapted using only the limited labeled subset $\mathcal{D}_l$ available to the student, requiring no additional annotations. Adaptation substantially improves performance across diverse domains, particularly where zero-shot transfer is poorly aligned with the target data.}
    \label{fig:teacher_adaptation}
    % \vspace{-0.6cm}
\end{figure}

\subsection{Overview}
\label{sec:method_overview}

We consider a target domain with a small labeled set
$\mathcal{D}_l=\{(x_i,y_i)\}_{i=1}^{N_l}$ and a substantially larger unlabeled
set $\mathcal{D}_u=\{x_j\}_{j=1}^{N_u}$, where $N_l \ll N_u$.
The deployed model is a compact real-time detector and is the only model used at
inference time. A vision foundation model (VFM) is available during training
only and serves as an offline annotation teacher.

Our pipeline consists of three stages, as illustrated in
\cref{fig:pipeline}. First, we adapt the VFM to the target domain using
$\mathcal{D}_l$. Second, we run the adapted teacher over $\mathcal{D}_u$ to
generate pseudo-annotations together with confidence estimates. Third, we
jointly train the compact student on $\mathcal{D}_l$ and the pseudo-labeled
$\mathcal{D}_u$ using reliability-aware pseudo-supervision. Rather than treating
teacher predictions with a single hard threshold, we use their confidence to
distinguish reliable supervision from uncertain regions and handle the latter
explicitly during training.

The VFM is used only for target adaptation and offline pseudo-annotation. Once
the pseudo-labeled training set has been constructed, the student is trained
independently and is the only model retained for deployment.

\subsection{Generating Pseudo-Labels}
\label{sec:method_pseudo_labels}

\noindent\textbf{Adapting the teacher.}
We use SAM~3~\cite{carion2026sam3} as the VFM teacher. Before generating
pseudo-labels, we fine-tune it on the labeled target set $\mathcal{D}_l$ using
standard supervised fine-tuning. These are the same annotations available to
the student, so teacher adaptation requires no additional labeling effort.

As shown in \cref{fig:teacher_adaptation}, the zero-shot teacher exhibits
substantial variation across target domains (details in \Cref{sec:exp_setup}), whereas adaptation consistently
improves its target-domain accuracy. The improvement is especially pronounced
when zero-shot transfer is weak, suggesting that limited target supervision can
effectively specialize the broad representation of the foundation model to the
deployment domain. This observation motivates a key design choice in our
pipeline: rather than using the VFM directly as a zero-shot annotator, we first
adapt the teacher and then amortize the resulting improvement over the much
larger unlabeled set. Additional fine-tuning details are provided in the
supplementary material.

\noindent\textbf{Pseudo-label construction.}
We run the adapted teacher once over $\mathcal{D}_u$ and record its predicted
classes, bounding boxes, and confidence scores. Duplicate predictions are
removed using non-maximum suppression. High-confidence detections provide
positive pseudo-labels, while lower-confidence detections are retained as
uncertain predictions rather than being immediately treated as background.
We describe how these uncertain predictions are handled during student
training in \cref{sec:method_loss}.

% \begin{table}[t]
%   \centering
%   \small
%   \caption{\textbf{Effect of teacher fine-tuning.} Teacher performance on the
%   target validation set of each dataset, before and after fine-tuning on
%   $\mathcal{D}_l$. P and R denote precision and recall; AP is averaged over IoU
%   thresholds $0.50\!:\!0.95$, and AP$_S$, AP$_M$, AP$_L$ report the same metric
%   restricted to small, medium, and large objects.}
%   \label{tab:teacher_adaptation}
%   \setlength{\tabcolsep}{4pt}
%   \resizebox{\columnwidth}{!}{%
%   \begin{tabular}{@{}llcccccc@{}}
%     \toprule
%     Dataset & Teacher & P & R & AP & AP$_S$ & AP$_M$ & AP$_L$ \\
%     \midrule
%     \multirow{2}{*}{\TODO{Dataset A}}
%       & Zero-shot   & -- & -- & -- & -- & -- & -- \\
%       & Fine-tuned  & -- & -- & -- & -- & -- & -- \\
%     \midrule
%     \multirow{2}{*}{\TODO{Dataset B}}
%       & Zero-shot   & -- & -- & -- & -- & -- & -- \\
%       & Fine-tuned  & -- & -- & -- & -- & -- & -- \\
%     \midrule
%     \multirow{2}{*}{\TODO{Dataset C}}
%       & Zero-shot   & -- & -- & -- & -- & -- & -- \\
%       & Fine-tuned  & -- & -- & -- & -- & -- & -- \\
%     \bottomrule
%   \end{tabular}}
% \end{table}

\subsection{Training the student}
\label{sec:method_loss}

The student is trained on $\mathcal{D}_l$ with ground-truth annotations and on
$\mathcal{D}_u$ with the pseudo-labels of \Cref{sec:method_pseudo_labels}. As in
prior semi-supervised detection
work~\cite{sohn2020stac,liu2021unbiased,xu2021softteacher}, we account for the
difference in reliability between the two sources with a scalar trade-off,
\begin{equation}
  \mathcal{L} = \mathcal{L}_s + \lambda_u \mathcal{L}_u,
  \label{eq:total-loss}
\end{equation}
where $\mathcal{L}_s$ and $\mathcal{L}_u$ are the losses on $\mathcal{D}_l$ and
$\mathcal{D}_u$ respectively, and $\lambda_u$ sets the overall strength of the
pseudo-supervision.

\noindent\textbf{Confidence-weighted loss.}
The scalar $\lambda_u$ expresses a single belief about the entire unlabeled set,
but pseudo-label quality is not uniform. The teacher is confident about some
objects and unsure about others, and it is often both within one image. We
therefore use the teacher's confidence in each pseudo-label as a proxy for its
quality and weight the label's contribution to the loss accordingly:
\begin{equation}
  \mathcal{L}_u = \frac{1}{Z} \sum_{k \in \mathcal{P}} w_k
  \Big[ \mathcal{L}_\text{cls}(\hat{c}_k, c_k)
      + \mathcal{L}_\text{box}(\hat{b}_k, b_k)
      + \mathcal{L}_\text{dfl}(\hat{b}_k, b_k) \Big].
  \label{eq:l-unsup}
\end{equation}

Here $\mathcal{P}$ is the set of student predictions that the detector's label
assignment matches to a pseudo-label, $c_k$ and $b_k$ are the class and box
targets supplied by the matched pseudo-label, $\hat{c}_k$ and $\hat{b}_k$ are
the student's predictions, $w_k$ is the teacher's confidence in the matched
pseudo-label, and $Z$ is the detector's usual per-batch normalizer.
$\mathcal{L}_s$ takes the same form with $w_k = 1$ for every ground-truth
target.

The three terms in \cref{eq:l-unsup} are the classification, localization, and
distribution focal terms of the detector we instantiate. The weighting is
independent of their form and applies to any detector whose loss decomposes over
individually assigned predictions.

\noindent\textbf{Zero-gating.}
While confidence filtering at $\tau$ removes detections that are unlikely to
correspond to an object, surviving detections can still be a real object for
which the teacher is uncertain in its class or localization. A common way to
handle these detections is to raise the threshold. STAC~\cite{sohn2020stac} and
Soft Teacher~\cite{xu2021softteacher} both use a high confidence threshold to
keep pseudo-labels precise. This is effective, but it also discards uncertain
detections of real objects. Once such a detection is discarded, the object is
unlabeled and its region can be assigned as background by the
student. The uncertainty of the teacher becomes a
negative example. We want to suppress the supervision from uncertain detections
without incurring this cost.

\begin{figure}[t]
  \centering
  % \fbox{\rule{0pt}{1.8in} \rule{0.9\linewidth}{0pt}}
  \includegraphics[width=\linewidth]{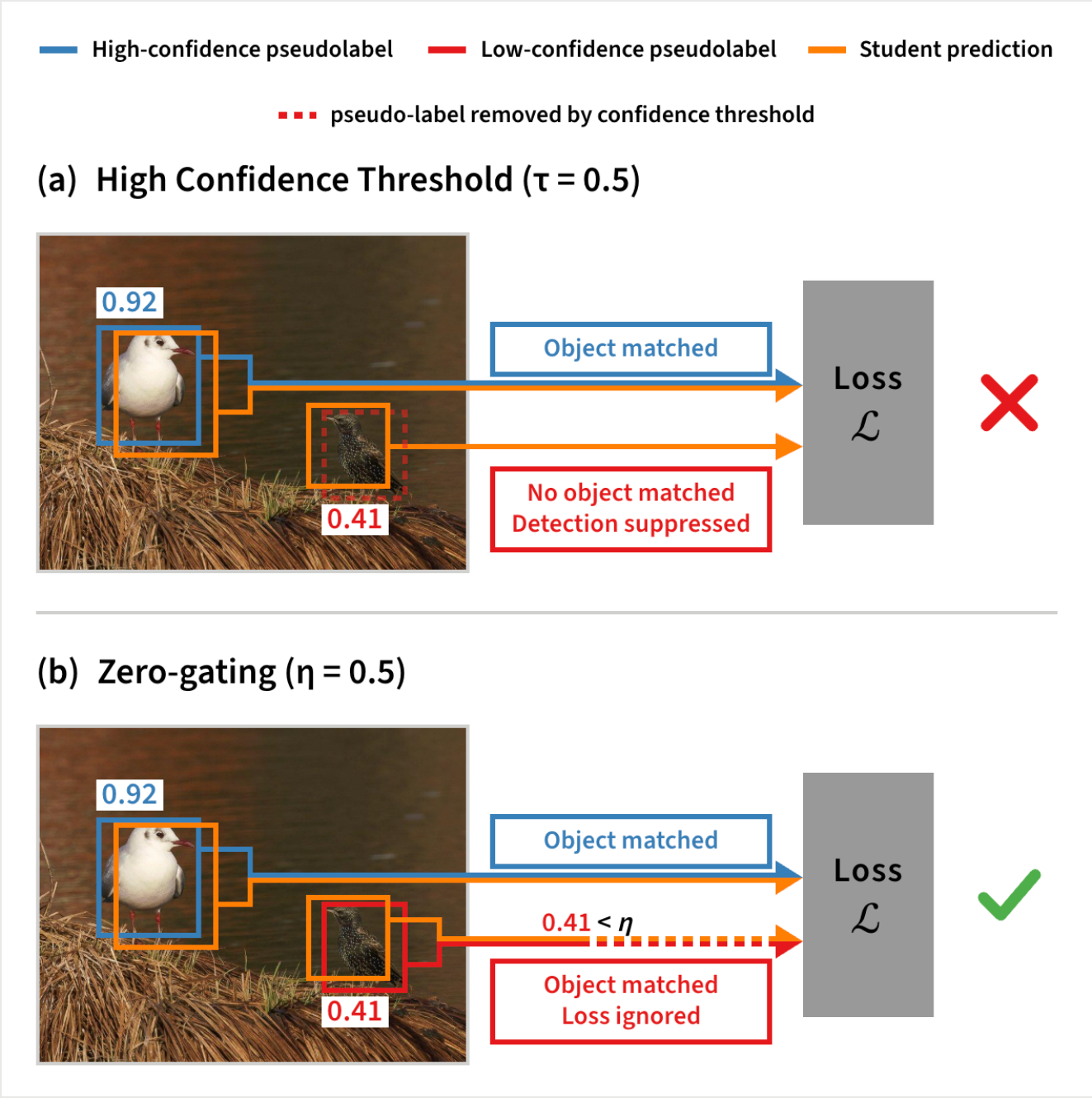}
  % \caption{\textbf{Zero-gating versus raising the confidence threshold.}
  % An uncertain pseudo-label on a real object is shown under both treatments.
  % Raising the threshold discards the pseudo-label, leaving the object
  % unlabeled, so label assignment treats its region as background and the student
  % learns to suppress it. Zero-gating retains the pseudo-label with weight zero:
  % the region is matched by the assigner and therefore not a negative, but it
  % contributes no gradient, and the image is left unmodified.}
  \caption{\textbf{Zero-gating versus high-threshold filtering.}
Thresholding discards uncertain pseudo-labels and may turn real objects into background negatives. Zero-gating keeps these hypotheses for assignment but sets their loss weight to zero.}
  \label{fig:zerogating}
  \vspace{-0.6cm}
\end{figure}

% We keep the detections produced by filtering and use a second threshold
% $\eta > \tau$ to decide whether they supervise the student. A pseudo-label with
% confidence below $\eta$ is assigned $w_k = 0$ in the confidence-weighted loss.
% It remains available to label assignment, so the object hypothesis is kept and
% its region is not matched as background, but it contributes no gradient. The
% object is ignored by the loss rather than mislabeled
% (\cref{fig:zerogating}).
We keep the detections produced by filtering and use a second threshold
$\eta > \tau$ to determine their contribution to the student loss.
Pseudo-labels with confidence below $\eta$ remain available for label
assignment but receive $w_k = 0$ in the confidence-weighted loss.
This suppresses loss contributions from student predictions assigned
to these uncertain pseudo-labels without modifying the input image
(\cref{fig:zerogating}).

% An alternative to zero-gating is to mask uncertain objects in the input image. Masking also prevents these regions from being learned as background, but it changes the input, removes contextual information, and introduces artifacts that are absent at deployment. Zero-gating leaves the image unchanged and operates only on the loss. 

% For retained pseudo-labels with confidence below $\eta$, we set their loss weights to zero while keeping them available for label assignment. This suppresses losses on predictions assigned to uncertain pseudo-labels without modifying the input image.
% We compare the two approaches in Supplementary material.

%% file: sec/4_experiments.tex
\section{Experiments}
\label{sec:experiments}

\subsection{Datasets and Evaluation Protocol}
\label{sec:exp_setup}

\noindent\textbf{Datasets.}
We evaluate on six target-domain object detection datasets spanning diverse
sources of deployment mismatch, including imaging modality, viewpoint and
scale, visual appearance, fine-grained category semantics, and
application-specific label definitions. Four datasets are drawn from
Roboflow100-VL~\cite{robicheaux2025roboflow100vl}:
\textit{Xray}, \textit{Pig Detection}, \textit{Fruitjes}, and
\textit{Floating Waste}, covering medical, agricultural, industrial, and
environmental domains, respectively. We additionally include
\textit{Garbage Classification 3}~\cite{garbage_classification_3}, an
application-specific waste-sorting dataset, and
\textit{VisDrone2019-DET}~\cite{du2019visdrone}, which provides a
complementary aerial setting with substantial viewpoint, scale, and
small-object shift. Dataset statistics are summarized in
\cref{tab:datasets}.

\begin{table}[t]
\centering
\small
\caption{\textbf{Target-domain datasets.}
The benchmarks span diverse application domains and sources of distribution
shift.}
\label{tab:datasets}
\setlength{\tabcolsep}{3.5pt}
\resizebox{\columnwidth}{!}{%
\begin{tabular}{@{}llcc@{}}
\toprule
\textbf{Dataset} & \textbf{Domain} & \textbf{Images} & \textbf{Classes} \\
\midrule
Xray                     & Medical       & 3,839  & 6  \\
Pig Detection            & Agriculture   & 6,148  & 4  \\
Fruitjes                 & Industrial    & 5,970  & 7  \\
Floating Waste           & Environmental & 3,031  & 5  \\
Garbage Classification 3 & Waste sorting & 10,464 & 7  \\
VisDrone2019-DET         & Aerial        & 8,629  & 10 \\
\bottomrule
\end{tabular}}
\vspace{-0.6cm}
\end{table}

\begin{table*}[t]
\centering
\small
\caption{\textbf{Detection performance under limited target-domain annotation.}
Student models use either full supervision or a $5\%$ labeled subset with
unlabeled target images; teacher performance is included for reference.
Results are mAP$_{50:95}$ on held-out test sets.
\textbf{Avg.} is computed over all domain datasets,
which are evaluated for all student sizes.}
\label{tab:main_result}

\setlength{\tabcolsep}{5pt}
\resizebox{\textwidth}{!}{%
\begin{tabular}{@{}lclccccccc@{}}
\toprule
\textbf{Training setting} &
\textbf{Human GT} &
\textbf{Model} &
\textbf{Garbage} &
\textbf{Fruitjes} &
\textbf{Floating Waste} &
\textbf{VisDrone} &
\textbf{Xray} &
\textbf{Pig Detection} &
\textbf{Avg.} \\
\midrule

Zero-shot teacher
& $0\%$ & SAM~3
& 31.2 & 11.1 & 18.8 & 26.1 & 0.1 & 2.4 & 14.9 \\

Adapted teacher
& $5\%$ & SAM~3
& 58.8 & 64.0 & 41.1 & 27.1 & 67.5 & 60.7 & 53.2 \\

\midrule

\oracle{\multirow{3}{*}{Full supervision}}
& \oracle{$100\%$} & \oracle{YOLOv11n}
& \oracle{53.0} & \oracle{68.4} & \oracle{46.0}
& \oracle{14.9} & \oracle{76.3} & \oracle{72.0}
& \oracle{55.1} \\

& \oracle{$100\%$} & \oracle{YOLOv11s}
& \oracle{54.2} & \oracle{69.3} & \oracle{48.4}
& \oracle{17.6} & \oracle{76.9} & \oracle{72.6}
& \oracle{56.5} \\

& \oracle{$100\%$} & \oracle{YOLOv11m}
& \oracle{54.1} & \oracle{69.9} & \oracle{50.3}
& \oracle{20.3} & \oracle{77.1} & \oracle{72.6}
& \oracle{57.3} \\

\midrule

\multirow{3}{*}{Labeled images only}
& $5\%$ & YOLOv11n
& 32.5 & 58.0 & 31.6 & 8.2 &72.7  & 61.0 &  44.0\\

& $5\%$ & YOLOv11s
& 33.7 & 59.4& 36.4 & 10.4 & 72.7 &60.9  &  45.6\\

& $5\%$ & YOLOv11m
& 30.8 & 60.4& 36.4 & 12.5 &  73.0& 60.8 & 45.7 \\

\midrule

\multirow{3}{*}{STAC-style}
& $5\%$ & YOLOv11n
& 37.9 & 56.0 & 34.5 & 9.7 & 73.1 & 62.0 & 45.5 ($\uparrow1.5$) \\

& $5\%$ & YOLOv11s
& 39.0 & 60.2 & 37.1 & 12.2 & 73.0 & 61.9 & 47.2 ($\uparrow1.6$) \\

& $5\%$ & YOLOv11m
& 36.3 & 60.1 & 38.9 & 14.1 & 73.3 & 61.7 & 47.4 ($\uparrow1.7$) \\

\midrule

\multirow{3}{*}{\textbf{Ours}}
& $5\%$ & YOLOv11n
& 50.5 & 66.1 & 38.8 & 14.2 & 69.7 & 61.8
& \textbf{50.2 ($\uparrow6.1$)} \\

& $5\%$ & YOLOv11s
& 51.2 & 65.0 & 38.2 & 16.8 & 69.7 & 61.6
& \textbf{50.4 ($\uparrow4.8$)} \\

& $5\%$ & YOLOv11m
& 51.6 & 68.1 & 40.0 & 18.5 & 69.8 & 61.7
& \textbf{51.6 ($\uparrow5.9$)} \\

\bottomrule
\vspace{-0.4cm}
\end{tabular}}
\end{table*}

\begin{table}[t]
  \centering
  \small
  \caption{\textbf{Roadmap of the proposed method on Garbage Classification.}
  Results use YOLOv11n with the same $5\%$ annotation budget throughout.
  Starting from STAC-style self-training, we successively replace the
  self-teacher with SAM~3, adapt the teacher to the target domain, and add
  confidence-aware student training. All variants have identical deployment cost.
The adapted-teacher baseline in this ablation uses the unweighted
variant of our confidence-aware training recipe, whereas
Tables~\ref{tab:et_reference} and~\ref{tab:teacher_comparison_garbage} use the native YOLO hard-label loss.}
  \label{tab:roadmap}
  \setlength{\tabcolsep}{3pt}
  \begin{tabular}{@{}lccccc@{}}
    \toprule
    Configuration & AP & AP$_S$ & AP$_M$ & AP$_L$ \\
    \midrule
    STAC-style baseline       & 37.92 & 14.42 & 39.24 & 63.01 \\
    + Zero-shot teacher       & 32.74&10.90&36.49&56.91\\
    + Adapted teacher       & 48.89 & 21.32 & 52.18 &76.75\\
    % - Balanced loading       & -- & -- & -- & -- \\
    + Conf.\ weighting       & 50.32&21.87&53.92 &78.07\\
    + Zero-gating (full)          & 50.47&21.94&53.64&79.25 \\
    \bottomrule
  \end{tabular}
  \vspace{-0.6cm}
\end{table}

\noindent\textbf{Data splits.}
For each dataset, we reserve disjoint validation and test sets that are never
used for teacher adaptation, pseudo-label generation, or student training.
Unless otherwise stated, all reported results are measured on the held-out test
set. The remaining training images are partitioned into a labeled subset
$\mathcal{D}_l$ and an unlabeled subset $\mathcal{D}_u$.

We construct $\mathcal{D}_l$ using a nominal $5\%$ annotation budget while
ensuring sufficient supervision for teacher adaptation. We first sample images
at random until every class is represented by at least $10$ labeled instances.
If this subset contains fewer than $5\%$ of the available training images, we
continue sampling uniformly at random until the $5\%$ budget is reached. If
satisfying the $10$-instance requirement already exceeds $5\%$, we retain the
resulting subset. All remaining training images constitute $\mathcal{D}_u$.
Details of teacher adaptation are provided in
\cref{sec:teacher_train}.

\noindent\textbf{Evaluation metrics.}
Our primary metric is mean average precision, averaged over IoU thresholds from
$0.50$ to $0.95$ (mAP$_{50:95}$). We additionally report AP$_{50}$ and,
where informative, scale-specific AP$_S$, AP$_M$, and AP$_L$. All detection
metrics are evaluated with a confidence threshold of $0$ and at most $300$
detections per image. Teacher and student models are evaluated on the same
held-out test split using the same metric definition. AP values are reported on
a $0$--$100$ scale.

\subsection{Implementation Details}
\label{sec:exp_impl}

\noindent\textbf{Teacher.}
We use SAM~3~\cite{carion2026sam3} as the foundation teacher. During
target-domain adaptation, the vision and language backbones are frozen, and
only the detection transformer head and class embeddings are optimized.
We train with AdamW for 20 epochs using an initial learning rate of
$8\times10^{-5}$, batch size 2, and $1008\times1008$ input resolution.
Training uses bfloat16 mixed precision. Full optimizer, schedule, loss, and
augmentation settings are provided in the supplementary material.

\noindent\textbf{Student.}
We use YOLOv11~\cite{khanam2024yolov11} as the compact detector and evaluate
the YOLOv11n, YOLOv11s, and YOLOv11m variants where indicated. Students are
trained for 100 epochs at $640\times640$ resolution with batch size 16 using
the Ultralytics auto-selected optimizer (MuSGD, initial learning rate $0.01$).
We otherwise retain the default YOLOv11 loss weights and training
augmentations. Full training settings are reported in the supplementary
material.

\noindent\textbf{Pseudo-label and confidence parameters.}
Unless otherwise stated, we use a pseudo-label retention threshold
$\tau=0.25$, NMS IoU threshold $\theta_{\mathrm{IoU}}=0.5$, and zero-gating
threshold $\eta=0.5$. The unlabeled loss weight is set to
$\lambda_u=0.5$. The same configuration is used across
datasets.

\noindent\textbf{Hardware.}
Teacher adaptation, offline pseudo-label generation, and student training are
performed on a single NVIDIA GeForce RTX 5090 GPU. Deployment efficiency is
measured separately on an NVIDIA Jetson AGX Orin, where we report per-image
latency and throughput in \cref{tab:efficiency}.

\subsection{Main Results}
\label{sec:exp_results}
\noindent\textbf{Main results.}
\Cref{tab:main_result} shows three trends across the six target domains.
First, adapting SAM~3 with the limited labeled subset substantially improves
teacher accuracy over zero-shot predictions across all six domains.
Second, this supervision yields sizeable gains over STAC-style self-training
on several datasets, bringing compact students closer to full supervision
despite the limited annotation budget.

Third, successful teacher adaptation does not guarantee positive transfer
to the student. On Xray, YOLOv11m achieves 69.8 AP compared with 73.3 AP
for STAC-style self-training; on Pig Detection, both achieve 61.7 AP.
In these domains, the adapted teacher remains below or close to the
limited-label student in accuracy. The limited transfer may reflect
difficulties in aligning the teacher with domain-specific label semantics,
including specialized anatomical categories in Xray and fine-grained
breed distinctions in Pig Detection. These distinctions may remain
challenging to learn from the small labeled subset despite substantial
improvements over zero-shot predictions. Overall, the results suggest
that the value of foundation-model supervision depends on the
task-specific knowledge it contributes beyond the limited-label student.
Aggregate AP alone, however, does not establish semantic misalignment
as the cause of the observed transfer limitations.

% \begin{table}[t]
%   \centering
%   \small
%   \caption{\textbf{Roadmap to the full method on \AW{TODO: PRIMARY DATASET}.} Each
%   row applies a single change to the row above it, starting from a STAC-style
%   self-training baseline in which the compact detector produces its own
%   pseudo-labels from $\mathcal{D}_l$. AP is reported on the held-out test set
%   under the protocol of \cref{sec:exp_data}. All configurations use the same
%   annotation budget and the same student architecture, so inference cost is
%   unchanged across rows.}
%   \label{tab:main}
%   \setlength{\tabcolsep}{3pt}
%   \begin{tabular}{@{}lccccc@{}}
%     \toprule
%     Configuration & AP & AP$_S$ & AP$_M$ & AP$_L$ \\
%     \midrule
%     STAC-style baseline       & 36.34 & 13.98 & 37.04 & 60.74 \\
%     + Zero-shot teacher       & -- & -- & -- & -- \\
%     + Adapted teacher       & -- & -- & -- & -- \\
%     % - Balanced loading       & -- & -- & -- & -- \\
%     + Conf.\ loss       & 51.13 & 22.88 & 54.92 & 79.42 \\
%     + Conf.\ masking (full)          & 51.64 & 22.52 & 55.52 & 79.74 \\
%     \bottomrule
%   \end{tabular}
% \end{table}

%% file: sec/5_analysis.tex
\subsection{Analysis}
\label{sec:analysis}

\noindent\textbf{What Makes Foundation-Model Supervision Effective?} \Cref{tab:roadmap} decomposes the proposed pipeline step by step using the same
annotation budget and student architecture throughout. Simply replacing the
self-trained teacher with zero-shot SAM~3 is not sufficient: AP decreases from
37.92 to 32.74, showing that foundation-model scale alone does not overcome
target-domain misalignment. In contrast, adapting the teacher with the limited
labeled subset raises AP to 48.89, accounting for the majority of the overall
improvement. This confirms that the main benefit comes from specializing the
foundation model to the target domain before using it as an annotator.

The remaining gap is due to imperfect pseudo-labels. Per-box confidence
weighting further improves AP from 48.89 to 50.32 by allowing reliable teacher
predictions to contribute more strongly while attenuating uncertain
supervision. Confidence weighting alone, however, does not address the
missing-label failure mode: if an uncertain object hypothesis is discarded, the
corresponding region may still be optimized as background. Zero-gating
addresses this complementary error by retaining such hypotheses during label
assignment while setting their loss weight to zero, yielding a further
improvement to 50.47 AP. 
% These regions therefore provide neither unreliable
% positive supervision nor unintended negative supervision.

Together, the roadmap suggests a simple progression: \emph{adaptation improves
what the teacher knows, confidence weighting controls how strongly that
knowledge is transferred, and zero-gating determines what should not be learned
when the teacher is uncertain}. Each component addresses a distinct failure
mode, and their combination gives the strongest student without changing its
inference architecture or cost.
Additional sensitivity analyses of $\lambda_u$ and $\eta$ are provided in the supplementary material.
\begin{figure*}[!t]
    \centering
    \vspace{-0.4cm}
    \includegraphics[width=\textwidth]{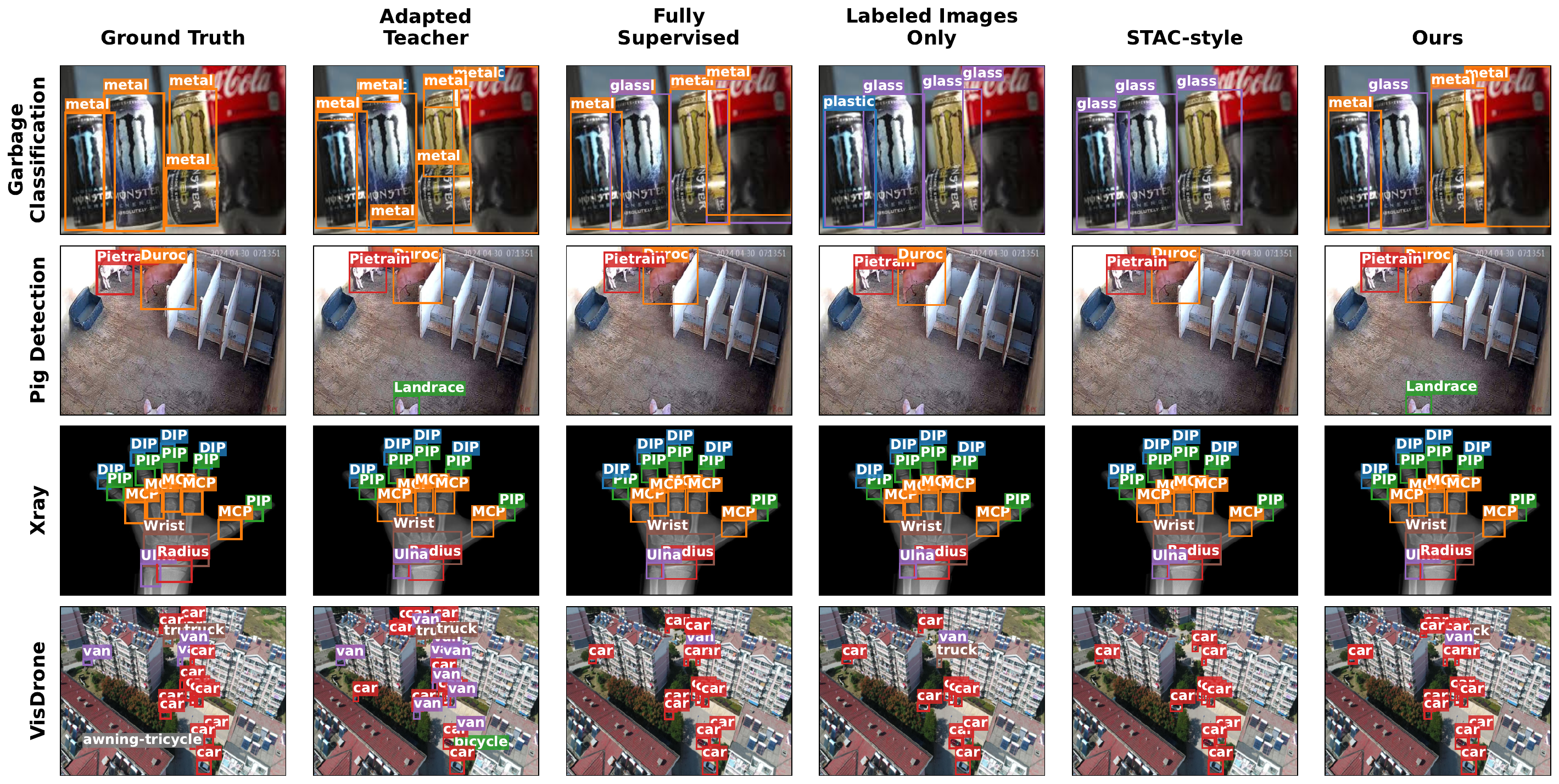}
    \vspace{-0.6cm}
    \caption{\textbf{Qualitative comparison across target domains.}
    Columns show ground truth, adapted SAM~3, fully supervised YOLOv11m,
    labeled-only training, STAC-style self-training, and ours. The examples
    illustrate semantic correction, improved small-object coverage, incomplete
    ground-truth annotations, and cases where the limited-label detector is
    already strong.}
    \label{fig:qual}
    % \vspace{-0.6cm}
\end{figure*}

\noindent\textbf{Comparison with Online SSOD}
\label{sec:online_ssod_comparison}
\input{tabs/et_reference}
% Table~\ref{tab:et_reference} compares our offline supervision with
% EfficientTeacher~\cite{xu2023efficientteacher} adapted to YOLOv11n on the
% same data splits. Our method achieves 13.19 and 2.04 AP higher than ET
% on Garbage and VisDrone, respectively. We also include GT-only continuation
% to distinguish online SSOD gains from additional supervised training.
% Protocol differences and control analyses are detailed in
% Appendix~\ref{sec:et_recovery}.
Table~\ref{tab:et_reference} compares adapted-SAM3 offline supervision with EfficientTeacher~\cite{xu2023efficientteacher} adapted to YOLOv11n on the same data splits. The hard-label SAM3 student achieves 13.19 and 2.04 AP higher than ET on Garbage and VisDrone, respectively. GT-only continuation controls for additional supervised training within the ET pipeline. These are method-level comparisons: optimization, augmentation, and teacher computation differ, so the AP gaps do not establish an equal-compute advantage. Further details are provided in Appendix~\ref{sec:et_recovery}.

\noindent\textbf{Effect of teacher choice.}
We evaluate three fine-tuned pseudo-labeling teachers on Garbage
using the same YOLOv11n student, data splits, and training protocol.
All students are trained with ground-truth annotations and hard
pseudo-labels, without confidence-weighted losses.

% \begin{table}[t]
%     \centering
%     \small
%     \caption{\textbf{Teacher comparison on Garbage.}
%     All teachers are fine-tuned; AP is reported for YOLOv11n.}
%     \label{tab:teacher_comparison_garbage}
%     \setlength{\tabcolsep}{5pt}
%     \begin{tabular}{@{}lr@{}}
%         \toprule
%         \textbf{Teacher} & \textbf{Student AP} \\
%         \midrule
%         None (GT only) & 32.54 \\
%         YOLOv11x~\cite{khanam2024yolov11} & 39.24 \\
%         Grounding DINO~\cite{liu2024groundingdino} & 43.22 \\
%         SAM3~\cite{carion2026sam3} & \textbf{47.95} \\
%         \bottomrule
%     \end{tabular}
%     % \vspace{-0.6cm}
% \end{table}

\begin{table}[t]
    \centering
    \small
    \caption{\textbf{Teacher comparison on Garbage.}
    All teachers are fine-tuned. Teacher and YOLOv11n student
    AP$_{50:95}$ are evaluated on the same test split.}
    \label{tab:teacher_comparison_garbage}
    \setlength{\tabcolsep}{4pt}
    \begin{tabular}{@{}lrr@{}}
        \toprule
        \textbf{Teacher} & \textbf{Teacher AP} & \textbf{Student AP} \\
        \midrule
        None (GT only) & -- & 32.54 \\
        YOLOv11x~\cite{khanam2024yolov11} & 35.13 & 39.24 \\
        Grounding DINO~\cite{liu2024groundingdino} & 44.62 & 43.22 \\
        SAM3~\cite{carion2026sam3} & \textbf{58.81} & \textbf{47.95} \\
        \bottomrule
    \end{tabular}
    \vspace{-0.6cm}
\end{table}

As shown in Table~\ref{tab:teacher_comparison_garbage},
all three teachers improve student AP over the GT-only baseline,
and the teacher ranking is preserved in the resulting students.
SAM3 achieves the highest teacher and student AP, with its student
reaching 47.95 AP: a gain of 15.41 points over GT-only training
and improvements of 4.73 and 8.71 points over Grounding DINO
and YOLOv11x supervision, respectively. Under the shared student
training protocol, higher teacher AP is thus accompanied by
better student performance across the three evaluated teachers.

The magnitude of the teacher advantage, however, is not fully
reflected in the student gap. SAM3 exceeds Grounding DINO by
14.19 teacher AP, while the corresponding students differ by
4.73 AP. Moreover, the YOLOv11x-supervised student achieves
39.24 AP, exceeding its teacher's 35.13 AP by 4.11 points.
Teacher AP therefore does not constitute a strict performance
ceiling for a student trained jointly on ground-truth annotations
and pseudo-labels. These results support SAM3 as the most effective
annotation teacher among the evaluated alternatives, with its
advantage realized through hard pseudo-labels alone, without
confidence-weighted losses.

\noindent\textbf{Efficiency.} Our framework is intentionally asymmetric: the foundation teacher is used only
to prepare pseudo-labels offline, whereas deployment requires only the compact
student. \Cref{tab:efficiency} quantifies this distinction on an NVIDIA Jetson AGX Orin. SAM3 requires 5992.6 GFLOPs per image and runs at only 2.9 FPS, making
continuous on-device inference impractical for real-time deployment. In
contrast, the YOLOv11 students are orders of magnitude cheaper. At
$1024\times1024$ resolution, YOLOv11n/s/m achieve 271, 164, and 80 FPS,
respectively; at $640\times640$, throughput increases to 600, 365, and 188 FPS.

The computational gap illustrates the role of the foundation model in our
pipeline. Rather than deploying its capacity directly, we use SAM3 to improve
the supervision available during training and transfer this benefit to a much
smaller detector. Teacher adaptation and pseudo-label generation are performed offline and do not affect the architecture, latency, or throughput of the deployed student.
% Consequently, all additional computation introduced by the
% teacher is incurred only during one-time pseudo-label generation and does not
% affect the architecture, latency, or throughput of the deployed student.

\begin{table}[t]
\centering
\small
\caption{\textbf{Edge-device inference efficiency on NVIDIA AGX Orin.}
The foundation teacher is used only offline for pseudo-label generation,
whereas YOLOv11 students are deployed for recurring inference.
GPU time and throughput are measured on the same Orin platform.
Slash-separated student values correspond to resolutions of 640/1024, respectively.}
\label{tab:efficiency}
\setlength{\tabcolsep}{4.5pt}
\resizebox{\columnwidth}{!}{%
\begin{tabular}{@{}llcccc@{}}
\toprule
\textbf{Role} & \textbf{Model} & \textbf{Resolution} & \textbf{GFLOPs} & \textbf{GPU time (ms)} & \textbf{FPS} \\
\midrule
Offline teacher & SAM3 & 1008 & 5992.6 & 345.02 & 2.9 \\
\midrule
\multirow{3}{*}{Deployed student}
& YOLOv11n & 640/1024 & 6.40/16.93 & 1.67/3.69 & 600/271 \\
& YOLOv11s & 640/1024 & 21.72/55.60 & 2.74/6.11 & 365/164 \\
& YOLOv11m & 640/1024 & 68.53/175.43 & 5.32/12.55 & 188/80 \\
\bottomrule
\end{tabular}}
\vspace{-0.6cm}
\end{table}
% \begin{table}[t]
%     \centering
%     \small
%     \caption{\textbf{Edge-device inference efficiency on NVIDIA AGX Orin.}
%     The foundation teacher is used only offline for pseudo-label generation,
%     whereas YOLOv11 students are deployed for recurring inference.
%     GPU time and throughput are measured on the same Orin platform.}
%     \label{tab:efficiency}
%     \setlength{\tabcolsep}{4.5pt}
%     \resizebox{\columnwidth}{!}{%
%     \begin{tabular}{@{}llcccc@{}}
%         \toprule
%         \textbf{Role} & \textbf{Model} & \textbf{Resolution} & \textbf{GFLOPs} & \textbf{GPU time (ms)} & \textbf{FPS} \\
%         \midrule
%         Offline teacher
%             & SAM3
%             & 1008
%             & 5992.6 
%             & 345.02
%             & 2.9 \\
%         \midrule
%         \multirow{5}{*}{Deployed student}
%         & YOLOv11n
%             & 640
%             & 6.40
%             & 1.67
%             & 600 \\
%         & YOLOv11s
%             & 640
%             & 21.72
%             & 2.74
%             & 365 \\
%         & YOLOv11m
%             & 640
%             & 68.53
%             & 5.32
%             & 188 \\
            
%             & YOLOv11n
%             & 1024
%             & 16.93
%             & 3.69
%             & 271 \\
%         & YOLOv11s
%             & 1024
%             & 55.60
%             & 6.11
%             & 164 \\
%         & YOLOv11m
%             & 1024
%             & 175.43
%             & 12.55
%             & 80 \\
        
%         \bottomrule
%     \end{tabular}}
%     \vspace{-0.6cm}
% \end{table}

\noindent\textbf{Qualitative Analysis.} \Cref{fig:qual} illustrates several characteristic behaviors across target
domains. In waste sorting, the adapted teacher better follows the target
category semantics than limited-label or STAC-style training, and this
improvement transfers to our compact student. In the dense aerial scene, the
teacher also provides broader small-object coverage, which is partially
inherited by the student.

The agricultural example further reveals incomplete ground truth: the adapted
teacher detects a visually valid pig instance that is not annotated. This
highlights the risk of treating uncertain or unmatched detections as
background and motivates zero-gating. In contrast, the X-ray example shows a
regime where the limited-label detector is already strong and external
supervision adds little. Overall, the examples reinforce that foundation-model
supervision is most useful when the adapted teacher provides complementary
target-domain knowledge.

% \begin{table}[t]
%   \centering
%   \small
%   \caption{\textbf{One-time preparation versus recurring deployment cost.}
%   Teacher figures cover a single offline pass over $\mathcal{D}_u$ during
%   pseudo-label generation and are incurred once; student figures are measured at
%   inference on a single image and are incurred for every deployed prediction.
%   Parameter counts and latency appear in \cref{tab:oracle}. Only the shaded row
%   reflects deployment cost.}
%   \label{tab:efficiency}
%   \setlength{\tabcolsep}{4pt}
%   \resizebox{\columnwidth}{!}{%
%   \begin{tabular}{@{}llccc@{}}
%     \toprule
%     Stage & Model & GFLOPs & Peak mem.\ (GB) & Time \\
%     \midrule
%     Offline (once)       & Teacher & -- & -- & -- \\
%     \colorours
%     Deployed (per image) & Student & -- & -- & -- \\
%     \bottomrule
%   \end{tabular}}
% \end{table}

%% file: tabs/et_reference.tex
\begin{table}[t]
\centering\small
\caption{
\textbf{Comparison with EfficientTeacher~\cite{xu2023efficientteacher} adapted to YOLO11n.} All methods use the same student architecture and labeled-data split. The adapted-SAM3 student uses hard pseudo-labels without confidence weighting or zero-gating. Results are held-out test AP ($\%$).}
\label{tab:et_reference}
\setlength{\tabcolsep}{5pt}
\begin{tabular*}{\columnwidth}{@{\extracolsep{\fill}}lrr@{}}
\toprule
\textbf{Training} & \textbf{Garbage} & \textbf{VisDrone} \\
\midrule
Supervised warmup & 32.54 & 8.91 \\
\quad + GT-only continuation & 35.07 & 9.62 \\
\quad + ET online SSOD & 34.76 & 10.81 \\
\midrule
Adapted SAM3 + hard pseudo-labels & \textbf{47.95} & \textbf{12.85} \\
\bottomrule
\vspace{-0.8cm}
\end{tabular*}
\end{table}

%% file: sec/6_conclusion.tex
\section{Conclusion}

This work demonstrates how limited target annotations can specialize an offline foundation teacher whose supervision is then propagated through a larger unlabeled set. Our adapt–label–distill framework combines this supervision with confidence-guided student training, yielding substantial gains in several evaluated domains while preserving compact-detector inference efficiency. In the Garbage ablation, teacher adaptation accounts for most of the improvement, with smaller additional gains from confidence-aware student training. These results support using foundation models as adaptable annotation teachers for resource-constrained detection: their capabilities improve training supervision, while their computational cost remains confined to offline adaptation and pseudo-label generation.

% \noindent\textbf{Limitations and outlook.} Transfer remains dependent on the quality and coverage of teacher predictions, and confidence-guided training does not guarantee gains in every domain. Future work could better identify when teacher supervision complements student predictions and reduce the offline cost of teacher adaptation and pseudo-label generation.
\noindent\textbf{Limitations and outlook.}
Transfer remains dependent on the quality and coverage of teacher
predictions, and confidence-guided training does not guarantee
gains in every domain. Confidence weighting and zero-gating
operate on retained teacher hypotheses; they do not directly
provide positive pseudo-supervision for objects that the teacher
fails to propose. Future work could better identify when external
supervision complements student predictions, address teacher
omissions, and reduce the offline cost of teacher adaptation
and pseudo-label generation.
\section{Acknowledgments}
% This work was supported by the National Science Foundation (NSF) and Center for Smart Streetscapes (CS3) under NSF Cooperative Agreement No. EEC-2133516. This work was also funded in part by the National Oceanic and Atmospheric Administration (NOAA) Sea Grant Program under Award No. NA24OARX417C0407.
This work was supported in part by the National Science Foundation (NSF) and Center for Smart Streetscapes (CS3)
under NSF Cooperative Agreement No. EEC-2133516, and 
by the National Oceanic and Atmospheric Administration (NOAA) Sea Grant Program under Award No. NA24OARX417C0407.

%% file: sec/X_supp.tex
\clearpage
\setcounter{figure}{0} % reset counter 
\setcounter{table}{0} % reset counter
\setcounter{section}{0} % reset counter
\setcounter{page}{1}
\renewcommand{\thesection}{A.\arabic{section}}
\maketitlesupplementary

\section{Additional Implementation details}

\subsection{Teacher Adaptation Details}
\label{sec:teacher_train}

We fine-tune SAM~3 by updating only the detection transformer head and class
embeddings while keeping the vision and language backbones frozen. Optimization
uses AdamW with an initial learning rate of $8\times10^{-5}$ for 20 epochs at
$1008\times1008$ resolution and batch size 2. Training uses bfloat16 mixed
precision and gradient clipping. We apply box-noise jittering to the prompt
boxes during adaptation. The complete optimization, scheduling, loss-weight,
and augmentation settings are reported in \cref{tab:sam3-hparams}.

\subsection{Student Training Details}
\label{sec:student_train}

YOLOv11 students are trained using the Ultralytics training pipeline. Unless
otherwise stated, YOLOv11n, YOLOv11s, and YOLOv11m use the same training
recipe. The auto-selected optimizer resolves to MuSGD, and each student is
trained for 100 epochs at $640\times640$ resolution with batch size 16 and
automatic mixed precision. We retain the default YOLOv11 loss weights and
training schedule. Full optimization settings are given in
\cref{tab:yolo-hparams}, and the augmentation policy is summarized in
\cref{tab:yolo-aug}.

\begin{table}[t]
\centering
\small
\caption{\textbf{SAM3 teacher adaptation hyperparameters.}}
\label{tab:sam3-hparams}
\setlength{\tabcolsep}{4.5pt}
\resizebox{\columnwidth}{!}{%
\begin{tabular}{@{}ll@{}}
\toprule
\textbf{Hyperparameter} & \textbf{Value} \\
\midrule
Trainable modules
    & Detection transformer head, class embeddings \\
Frozen modules
    & Vision and language backbones \\
Optimizer
    & AdamW \\
Initial learning rate
    & $8\times10^{-5}$ \\
LR schedule
    & Inverse square root \\
Warmup / cooldown
    & 20 / 20 steps \\
Weight decay
    & $0.1$ \\
Batch size
    & 2 \\
Gradient accumulation
    & 1 \\
Training epochs
    & 20 \\
Samples per epoch
    & 1,500 \\
Total optimization steps
    & 15,000 \\
Input resolution
    & $1008\times1008$ \\
Precision
    & bfloat16 AMP \\
Gradient clipping
    & $L_2$ max norm $0.1$ \\
Box-noise jitter
    & $\sigma=0.1$, max 20 px \\
$\lambda_{\mathrm{bbox}}$
    & 5.0 \\
$\lambda_{\mathrm{giou}}$
    & 2.0 \\
$\lambda_{\mathrm{ce}}$
    & 20.0 \\
$\lambda_{\mathrm{presence}}$
    & 20.0 \\
Focal loss $(\alpha,\gamma)$
    & $(0.25,\,2.0)$ \\
Positive class weight
    & 10.0 \\
Seed
    & 123 \\
\bottomrule
\end{tabular}}
\end{table}
\begin{table}
\centering
\small
\caption{\textbf{YOLO11 student training hyperparameters.}
The same training recipe is used for YOLO11n, YOLO11s, and YOLO11m unless
otherwise stated.}
\label{tab:yolo-hparams}
\setlength{\tabcolsep}{4.5pt}
\begin{tabular}{@{}ll@{}}
\toprule
\textbf{Hyperparameter} & \textbf{Value} \\
\midrule
Optimizer
    & MuSGD (Ultralytics auto-selected) \\
Initial learning rate
    & $0.01$ \\
Final LR factor (\texttt{lrf})
    & $0.01$ \\
LR schedule
    & Linear decay \\
Warmup
    & 3 epochs \\
Warmup momentum
    & $0.8$ \\
Warmup bias LR
    & $0.1$ \\
Momentum
    & $0.9$ \\
Weight decay
    & $5\times10^{-4}$ \\
Batch size
    & 16 \\
Training epochs
    & 100 \\
Input resolution
    & $640\times640$ \\
Precision
    & AMP \\
$\lambda_{\mathrm{box}}$
    & 7.5 \\
$\lambda_{\mathrm{cls}}$
    & 0.5 \\
$\lambda_{\mathrm{dfl}}$
    & 1.5 \\
Seed
    & 42 \\
\bottomrule
\end{tabular}
\end{table}
\begin{table}
\centering
\small
\caption{\textbf{YOLO11 student augmentation policy.}}
\label{tab:yolo-aug}
\setlength{\tabcolsep}{5pt}
\begin{tabular}{@{}ll@{}}
\toprule
\textbf{Augmentation} & \textbf{Setting} \\
\midrule
Mosaic
    & $1.0$; disabled for final 10 epochs \\
HSV (hue / saturation / value)
    & $0.015 / 0.7 / 0.4$ \\
Translation
    & $0.1$ \\
Scale
    & $0.5$ \\
Horizontal flip
    & $0.5$ \\
Vertical flip
    & $0.0$ \\
Rotation / shear / perspective
    & $0.0 / 0.0 / 0.0$ \\
MixUp / CutMix / Copy-Paste
    & $0.0 / 0.0 / 0.0$ \\
Random erasing
    & $0.4$ \\
RandAugment
    & Enabled \\
\bottomrule
\end{tabular}
\end{table}

\section{Sensitivity Analysis}
\label{sec:supp_sensitivity}

\noindent\textbf{Joint sensitivity to pseudo-label strength and zero-gating.}
We jointly vary the unlabeled loss weight $\lambda_u$ and the zero-gating
threshold $\eta$ while fixing the initial pseudo-label retention threshold at
$\tau=0.25$. All remaining training and pseudo-labeling settings are held
fixed. For clarity, the reported main experiments use
$(\lambda_u,\eta)=(0.5,0.5)$.

\begin{figure*}[t]
    \centering
    \includegraphics[width=0.68\textwidth]{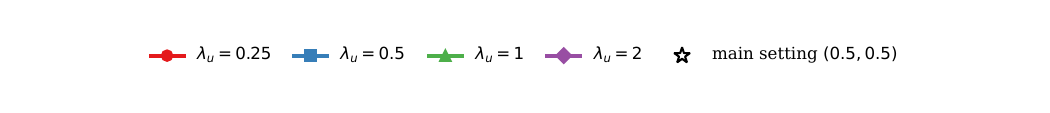}\\[-0.4em]
    \includegraphics[width=0.315\textwidth]{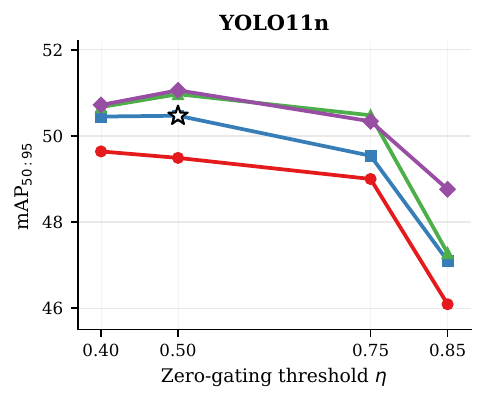}
    \hfill
    \includegraphics[width=0.315\textwidth]{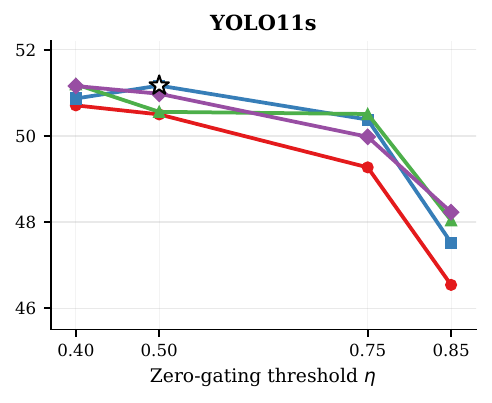}
    \hfill
    \includegraphics[width=0.315\textwidth]{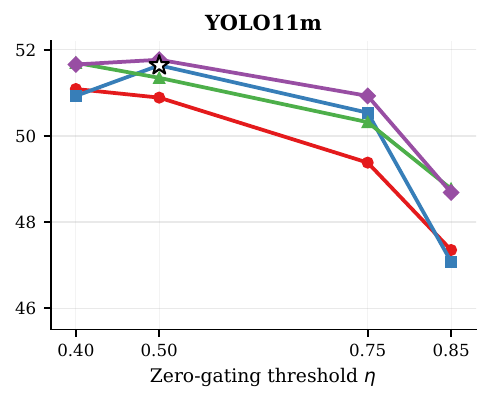}
    \caption{\textbf{Joint sensitivity to the unlabeled loss weight
    $\lambda_u$ and zero-gating threshold $\eta$.}
    Results are mAP$_{50:95}$ on Garbage Classification for three student
    capacities. The initial retention threshold is fixed at $\tau=0.25$.
    The black star marks the shared configuration used in the main experiments,
    $(\lambda_u,\eta)=(0.5,0.5)$.}
    \label{fig:sensitivity_lambda_eta}
\end{figure*}

\Cref{fig:sensitivity_lambda_eta} shows a broad stable regime across student
capacities. For $\lambda_u\in[0.5,2.0]$ and
$\eta\in\{0.4,0.5\}$, performance varies by only $0.61$, $0.64$, and
$0.84$ AP for YOLO11n, YOLO11s, and YOLO11m, respectively. The shared
configuration used in the main experiments is within $0.59$, $0.03$, and
$0.13$ AP of the best observed result for the three models. Thus, the method
does not require model-specific tuning or a sharply selected optimum.

Performance is more sensitive to $\eta$ than to $\lambda_u$. With the selected
$\lambda_u=0.5$, increasing $\eta$ from $0.5$ to $0.75$ causes only a modest
drop, whereas $\eta=0.85$ reduces performance by $3.37$, $3.65$, and
$4.56$ AP for YOLO11n, YOLO11s, and YOLO11m. This behavior is consistent
with overly aggressive zero-gating: too many teacher hypotheses are prevented
from contributing useful supervision. Increasing $\lambda_u$ can partially
strengthen the remaining pseudo-labels, but does not recover the stable regime.
Overall, a single moderate setting transfers reliably across student sizes.
\subsection{EfficientTeacher Baseline and Supervised Control}
\label{sec:et_recovery}

\paragraph{YOLO11n adaptation.}
We adapt EfficientTeacher~\cite{xu2023efficientteacher} to YOLO11n to
compare online and offline supervision with the same student architecture.
The adaptation retains teacher EMA updates, label matching, and geometric
alignment between augmented views. It uses YOLO11n's native task-aligned
assignment and distribution focal loss. Because YOLO11n has no separate
objectness branch, uncertain pseudo-labels receive a soft joint-class loss
without a localization loss. This is an adaptation of the method, rather
than a reproduction of its original YOLOv5 implementation.

\paragraph{Training and evaluation.}
We evaluate on Garbage and VisDrone using the same labeled and unlabeled
splits as the offline comparison. ET training consists of 100 supervised
warmup epochs followed by 100 online SSOD epochs. To measure the contribution
of online SSOD beyond longer supervised training, we branch a GT-only
continuation from the same warmup checkpoint. This control reuses only the
original human annotations and matches the planned update opportunities and
labeled-image presentations of the SSOD stage. It preserves the SGD schedule,
gradient accumulation, clipping threshold of 10, and dual-EMA validation
procedure, but omits the unlabeled forward pass. Thus, the control matches
the planned supervised training exposure, not total computation.
All runs use seed 42 and 640-pixel inputs. Checkpoints are selected on
validation data and scored with the shared held-out test evaluator.

\paragraph{Results and interpretation.}
On Garbage, online SSOD improves over warmup from 32.54 to 34.76 AP,
but GT-only continuation reaches 35.07 AP. The warmup comparison therefore
does not establish a benefit from unlabeled data in this run. On VisDrone,
online SSOD reaches 10.81 AP, compared with 8.91 AP after warmup and
9.62 AP after GT-only continuation, yielding a 1.19 AP advantage over the
latter. Adapted-SAM3 supervision reaches 47.95 and
12.85 AP, exceeding ET by 13.19 and 2.04 AP, respectively
(Table~\ref{tab:et_reference}). These single-seed comparisons show higher
student AP with our offline supervision in both domains. Differences in
optimization, augmentation, pretraining, and teacher cost prevent attributing
the gap solely to pseudo-label quality or claiming an equal-compute advantage.